\documentclass[11pt,a4paper]{article}
\usepackage[hyperref]{acl2020}
\usepackage{times}
\usepackage{latexsym}
\usepackage{framed}
\usepackage{graphicx}
\usepackage{caption}
\usepackage{subcaption}
\usepackage{dsfont}
\usepackage{xspace}
\usepackage{booktabs}  
\usepackage{amsmath}
\usepackage{todonotes}
\usepackage[normalem]{ulem}

\newcommand{\bertinterpret}[2]{\ensuremath{\mathcal{I}(#1,#2)}}

\DeclareMathOperator*{\mlp}{\ensuremath{\text{MLP}}}
\DeclareMathOperator*{\exbert}{\text{ExpBERT}}
\DeclareMathOperator*{\noexp}{\text{NoExp}}
\DeclareMathOperator*{\bertsemparse}{\text{BERT+SemParser}}
\DeclareMathOperator*{\exbertext}{\text{ExpBERT (+ External)}}

\DeclareMathOperator*{\bertregex}{\text{BERT+Patterns}}
\DeclareMathOperator*{\bert}{\ensuremath{\text{BERT}}}
\DeclareMathOperator*{\exbertprob}{\ensuremath{\text{ExpBERT-Prob}}}

\newcommand{\norm}[1]{\left\vert#1\right\vert}
\newcommand{\labelspace}{\mathcal{Y}}
\newcommand{\model}[1]{\ensuremath{f_\theta(#1)}}
\newcommand{\expl}[1]{\ensuremath{e_{#1}}}
\newcommand{\relexp}[1]{\ensuremath{r_{#1}}}
\newcommand{\repr}[1]{\ensuremath{u(#1)}}
\newcommand{\expvec}[1]{\ensuremath{v(#1)}}

\newcommand{\seq}{s}
\newcommand{\inp}{x}
\newcommand{\outp}{y}

\newcommand{\lmtt}[1]{\fontfamily{lmtt}\selectfont{#1}}

\newcommand{\spouse}{{\lmtt Spouse}\xspace}
\newcommand{\disease}{{\lmtt Disease}\xspace}
\newcommand{\tacred}{{\lmtt TACRED}\xspace}

\newcommand\sE{\ensuremath{\mathcal{E}}}

\newcommand\sI{\ensuremath{\mathcal{I}}}

\newcommand\sY{\ensuremath{\mathcal{Y}}}

\newcommand\bR{\ensuremath{\mathbf{R}}}

\newcommand\reffig[1]{Figure~\ref{fig:#1}}

\newcommand\reftab[1]{Table~\ref{tab:#1}}
\newcommand\refapp[1]{Appendix~\ref{sec:#1}}

\ifthenelse{\isundefined{\definition}}{\newtheorem{definition}{Definition}}{}
\ifthenelse{\isundefined{\assumption}}{\newtheorem{assumption}{Assumption}}{}
\ifthenelse{\isundefined{\hypothesis}}{\newtheorem{hypothesis}{Hypothesis}}{}
\ifthenelse{\isundefined{\proposition}}{\newtheorem{proposition}{Proposition}}{}
\ifthenelse{\isundefined{\theorem}}{\newtheorem{theorem}{Theorem}}{}
\ifthenelse{\isundefined{\lemma}}{\newtheorem{lemma}{Lemma}}{}
\ifthenelse{\isundefined{\corollary}}{\newtheorem{corollary}{Corollary}}{}
\ifthenelse{\isundefined{\alg}}{\newtheorem{alg}{Algorithm}}{}
\ifthenelse{\isundefined{\example}}{\newtheorem{example}{Example}}{}

\usepackage{microtype}

\aclfinalcopy 
\def\aclpaperid{1256} 

\title{ExpBERT: Representation Engineering with \\ Natural Language Explanations}
\author{Shikhar Murty \qquad Pang Wei Koh \qquad Percy Liang \\
Computer Science Department, Stanford University \\
\texttt{\{smurty,pangwei,pliang\}@cs.stanford.edu}
  }

\date{}

\begin{document}
\maketitle
\begin{abstract}
Suppose we want to specify the inductive bias that married couples typically go on honeymoons
for the task of extracting pairs of spouses from text. 
In this paper, we allow model developers to specify
these types of inductive biases as natural language explanations.
We use BERT fine-tuned on MultiNLI to ``interpret'' these explanations with respect to the input sentence,
producing explanation-guided representations of the input.
Across three relation extraction tasks, our method, $\exbert$, matches a BERT baseline but with 3--20$\times$ less labeled data and improves on the baseline by 3--10 F1 points with the same amount of labeled data.
\end{abstract}

\section{Introduction}

Consider the relation extraction task of finding spouses in text,
and suppose we wanted to specify the inductive bias that married couples typically go on honeymoons. In a traditional feature engineering approach, we might try to construct a ``did they go on a honeymoon?'' feature and add that to the model. 
In a modern neural network setting, however, it is not obvious how to use standard approaches like careful neural architecture design or data augmentation to induce such an inductive bias.
In a way, while the shift from feature engineering towards end-to-end neural networks and representation learning has alleviated the burden of manual feature engineering and increased model expressivity, it has also reduced our control over the inductive biases of a model.

\begin{figure}
\centering
\includegraphics[width=\linewidth]{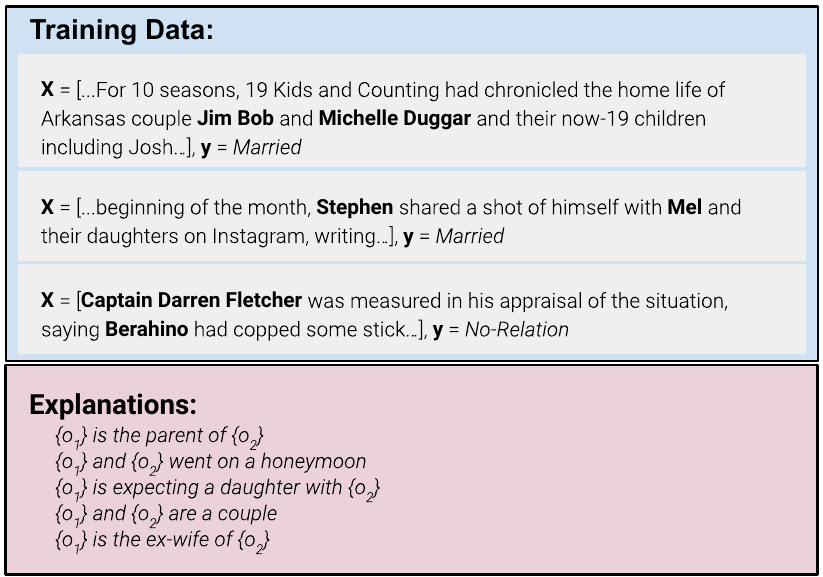}
\caption{Sample data points and explanations from \spouse, one of our relation extraction tasks. The explanations provide relevant features for classification.}
\vspace{-2mm}
\label{fig:exp_and_data}
\end{figure}

In this paper, we explore using natural language explanations (\reffig{exp_and_data})
to generate features that can augment modern neural representations. This imbues representations with inductive biases corresponding to the explanations, thereby restoring some degree of control while maintaining their expressive power.

Prior work on training models with explanations use semantic parsers to interpret explanations: 
the parser converts each explanation into an executable logical form that is executable over the input sentence 
and uses the resulting outputs as features \citep{srivastava-etal-2017-joint} or as noisy labels on unlabeled data \cite{Hancock2018TrainingCW}.
However, semantic parsers can typically only parse low-level statements like ``\textit{`wife' appears between \{$o_1$\} and \{$o_2$\} and the last word of \{$o_1$\} is the same as the last word of \{$o_2$\}}'' \citep{Hancock2018TrainingCW}.

\begin{figure}
\centering
\includegraphics[width=\linewidth]{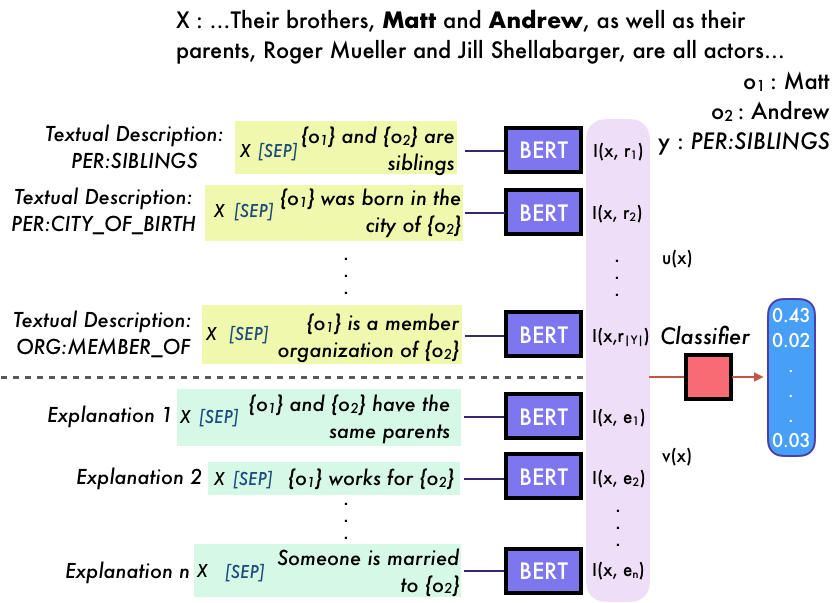}
\caption{Overview of our approach. Explanations as well as textual descriptions of relations are interpreted using $\bert$ for a given $\inp$ to produce a representation which form inputs to our classifier.}
\label{fig:expbert_overview}
\end{figure}

We remove these limitations
by using modern distributed language representations, 
instead of semantic parsers, 
to interpret language explanations. 
Our approach, $\exbert$ (Figure \ref{fig:expbert_overview}), uses $\bert$ \citep{Devlin2019BERTPO} fine-tuned on the MultiNLI natural language inference dataset \citep{multinli18} to produce features that ``interpret'' each explanation on an input.
We then use these features to augment the input representation.
Just as a semantic parser grounds an explanation by converting it into a logical form and then executing it,
the features produced by $\bert$ can be seen as a soft ``execution'' of the explanation on the input.

On three benchmark relation extraction tasks, $\exbert$ improves over a BERT baseline with no explanations: it achieves an F1 score of 3--10 points higher with the same amount of labeled data, and a similar F1 score as the full-data baseline but with 3--20x less labeled data. $\exbert$ also improves on a semantic parsing baseline (+3 to 5 points F1), suggesting that natural language explanations can be richer than low-level, programmatic explanations.

\section{Setup}
\paragraph{Problem.} We consider the task of relation extraction: Given $x = (\seq, o_1, o_2)$, where $\seq$ is a sequence of words and $o_1$ and $o_2$ are two entities that are substrings within $\seq$, our goal is to classify the relation $\outp \in \labelspace$ between $o_1$ and $o_2$. The label space $\labelspace$ includes a \textsc{No-Relation} label if no relation applies.
Additionally, we are given a set of natural language explanations $\mathcal{E} =  \{\expl{1}, \expl{2}, \ldots, \expl{n}\}$ designed to capture relevant features of the input for classification. These explanations are used to define a \emph{global} collection of features and are not tied to individual examples. 

\paragraph{Approach.} Our approach (Figure \ref{fig:expbert_overview}) uses pre-trained neural models to interpret the explanations $\mathcal{E}$ in the context of a given input $\inp$. 
Formally, we define an \emph{interpreter} $\sI$ as any function that takes an input $x$ and explanation $e_j$ and produces a feature vector in $\bR^d$.
In our $\exbert$ implementation, 
we choose $\sI$ to capture whether the explanation $e_j$ is entailed by the input $x$.
Concretely, we use $\bert$ \citep{Devlin2019BERTPO} fine-tuned on MultiNLI \citep{multinli18}: we feed wordpiece-tokenized versions of the explanation $\expl{j}$ (hypothesis) and the instance $\inp$ (premise), separated by a \textsc{[SEP]} token, to BERT. Following standard practice, we use the vector at the \textsc{[CLS]} token to represent the entire input as a 768-dimensional feature vector:
\begin{align}
&\bertinterpret{\inp}{\expl{j}} = \bert\big(\small{\text{[CLS]}}, \seq, \small{\text{[SEP]}}, \expl{j}\big).
\end{align}
These vectors, one for each of the $n$ explanations, are concatenated to form the \emph{explanation representation} $v(x) \in \bR^{768n}$,
\begin{align}
&\expvec{x} = \big[\bertinterpret{x}{\expl{1}}, \bertinterpret{x}{\expl{2}}, \ldots, \bertinterpret{x}{\expl{n}}\big].
\end{align}
In addition to $\expvec{x}$, we also map $\inp$ into an \emph{input representation} $\repr{x} \in \bR^{768|\sY|}$ by using the same interpreter over textual descriptions of each potential relation.
Specifically, we map each potential relation $y_i$ in the label space $\labelspace$ to a textual description $\relexp{i}$ (Figure \ref{fig:expbert_overview}), apply $\bertinterpret{x}{\cdot}$ to \relexp{i}, and concatenate the resulting feature vectors: 
\begin{align}
\repr{x} &= \big[\bertinterpret{\inp}{\relexp{1}}, \bertinterpret{\inp}{\relexp{2}}, \ldots, \bertinterpret{\inp}{\relexp{\norm{\labelspace}}}\big].
\end{align}
Finally, we train a classifier over $\repr{x}$ and $\expvec{x}$:
\begin{align}
 \model{x} &= \mlp \big[ \repr{x}, \expvec{x} \big].
\end{align}

Note that \repr{x} and \expvec{x} can be obtained in a pre-processing step since $\bertinterpret{\cdot}{\cdot}$ is fixed (i.e., we do not additionally fine-tune BERT on our tasks).
For more model details, please refer to \refapp{app-model}.

\paragraph{Baselines.} 
We compare $\exbert$ against several baselines that train a classifier over the same input representation $\repr{x}$. \textbf{$\noexp$} trains a classifier only on $\repr{x}$. 
The other baselines augment $\repr{x}$ with variants of the explanation representation $v(x)$.
\textbf{$\bertsemparse$} uses the semantic parser from \citet{Hancock2018TrainingCW} to convert explanations into executable logical forms. The resulting denotations over the input $x$ (a single bit for each explanation) are used as the explanation representation, i.e., $v(x) \in \{0, 1\}^n$. We use two different sets of explanations for this baseline: our natural language explanations (LangExp) and the low-level explanations from \citet{Hancock2018TrainingCW} that are more suitable for the semantic parser (ProgExp). \textbf{$\bertregex$} converts explanations into a collection of unigram, bigram, and trigram patterns and creates a binary feature for each pattern based on whether it is contained in $\seq$ or not. This gives $v(x) \in \{0, 1\}^{n'}$, where $n'$ is the number of patterns. Finally, we compare $\exbert$ against a variant called \textbf{$\exbertprob$}, where we directly use entailment probabilities obtained by $\bert$ (instead of the feature vector at the [CLS] token) as the explanation representation $v(x) \in [0,1]^n$.

\section{Experiments}
\paragraph{Datasets.} We consider 3 relation extraction datasets from various domains---\spouse and \disease \cite{Hancock2018TrainingCW}, and \tacred \cite{zhang2017tacred}. \spouse involves classifying if two entities are married; \disease involves classifying whether the first entity (a chemical) is a cause of the second entity (a disease); and \tacred involves classifying the relation between the two entities into one of 41 categories. Dataset statistics are in Table \ref{tab:data}; for more details, see \refapp{app-datasets}.

\paragraph{Explanations.}
To construct explanations, we randomly sampled 50 training examples for each $\outp \in \labelspace$ and wrote a collection of natural language statements explaining the gold label for each example.
For \spouse and \disease, we additionally wrote some negative explanations for the \textsc{No-Relation} category.
To interpret explanations for \disease, we use SciBERT, a variant of BERT that is better suited for scientific text \citep{Beltagy2019SciBERT}.
A list of explanations can be found in \refapp{app-explanations}. 

\begin{table}[]
\caption{Dataset statistics.}
\label{tab:data}
\centering
\small
\begin{tabular}{lrrrr} \toprule
Dataset & Train & Val   & Test & Explanations  \\ \hline
\spouse  & 22055 & 2784  & 2680 & 40  \\
\disease & 6667  & 773   & 4101 & 28 \\
\tacred  & 68124 & 22631 & 15509 & 128 \\ \bottomrule
\end{tabular}
\end{table}

\begin{table*}[]
\caption{Results on relation extraction datasets. For \spouse and \disease, we report 95\% confidence intervals and for \tacred, we follow the evaluation protocol from \citet{zhang2017tacred}. More details in Appendix \ref{sec:appendix}.}
\label{tab:results}
\centering
\small
\begin{tabular}{lrrr} \toprule
Model      & \spouse & \disease  & \tacred  \\ \hline
$\noexp$       & 52.9 $\pm$ 0.97 & 49.7 $\pm$ 1.01  &      64.7 \\
$\bertregex$ & 53.3 $\pm$ 1.24 & 49.0 $\pm$ 1.15    &   64.4    \\
$\bert$+SemParse (LangExp)    &  53.6 $\pm$ 0.38  & 49.1 $\pm$ 0.47                 &  -     \\
$\bert$+SemParse (ProgExp)    &  58.3 $\pm$ 1.10  & 49.7 $\pm$ 0.54  &    -   \\
$\exbertprob$ & 58.4 $\pm$ 1.22 & 49.7 $\pm$ 1.21 & 65.3 \\ 
$\exbert$     & \textbf{63.5 $\pm$ 1.40} & \textbf{52.4 $\pm$ 1.23} & \textbf{67.9}   \\
\bottomrule
\end{tabular}
\end{table*}

\paragraph{Benchmarks.} 
We find that explanations improve model performance across all three datasets:
$\exbert$ improves on the $\noexp$ baseline by +10.6 F1 points on \spouse, +2.7 points on \disease, and +3.2 points on \tacred (\reftab{results}).\footnote{We measure performance using F1 scores due to the class imbalance in the datasets (\spouse: 8\% positive, \disease: 20.8\% positive, and \tacred: 20.5\% examples with a relation).}
On \tacred, which is the most well-established of our benchmarks and on which there is significant prior work, $\exbert$ (which uses a smaller BERT-base model that is not fine-tuned on our task) outperforms the standard, fine-tuned BERT-large model by +1.5 F1 points \citep{joshi2019spanbert}. 
Prior work on \spouse and \disease used a simple logistic classifier over traditional features created from dependency paths of the input sentence. This performs poorly compared to neural models, and our models attain significantly higher accuracies \citep{Hancock2018TrainingCW}.

Using BERT to interpret natural language explanations improves on using semantic parsers to evaluate programmatic explanations (+5.5 and +2.7 over $\bertsemparse$ (ProgExp) on \spouse and \disease, respectively). $\exbert$ also outperforms the $\bertsemparse$ (LangExp) model by +9.9 and +3.3 points on \spouse and \disease. We exclude these results on \tacred as it was not studied in \citet{Hancock2018TrainingCW}, so we did not have a corresponding semantic parser and set of programmatic explanations.

We note that $\exbert$---which uses the full 768-dimensional feature vector from each explanation---outperforms ExpBERT (Prob), which summarizes these vectors into one number per explanation, by +2--5 F1 points across all three datasets. 

\begin{figure}
\centering
\begin{subfigure}{.49\linewidth}
  \centering
  \includegraphics[width=\linewidth]{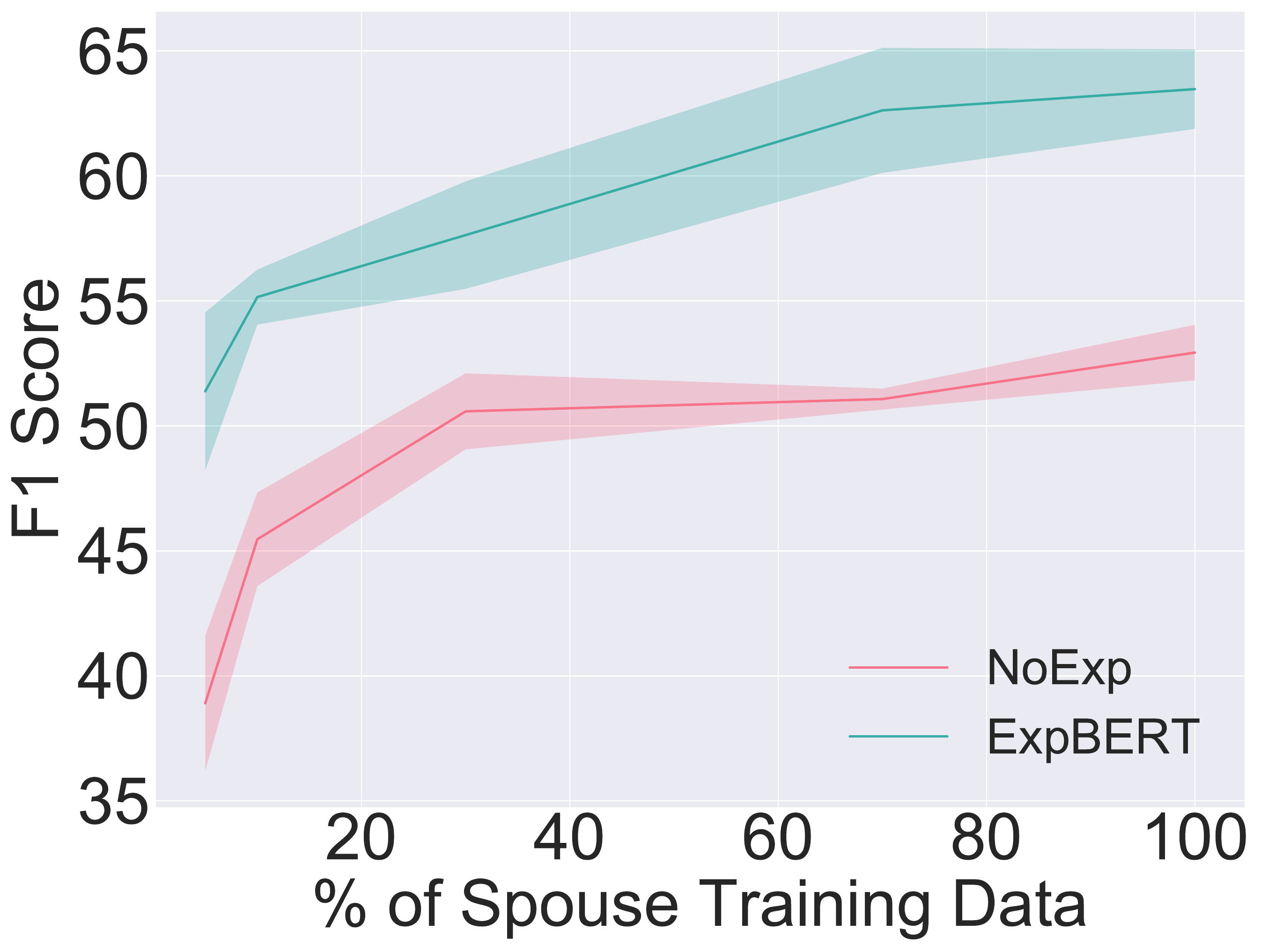}
  \label{fig:sub1}
\end{subfigure}
\begin{subfigure}{.49\linewidth}
  \centering
  \includegraphics[width=\linewidth]{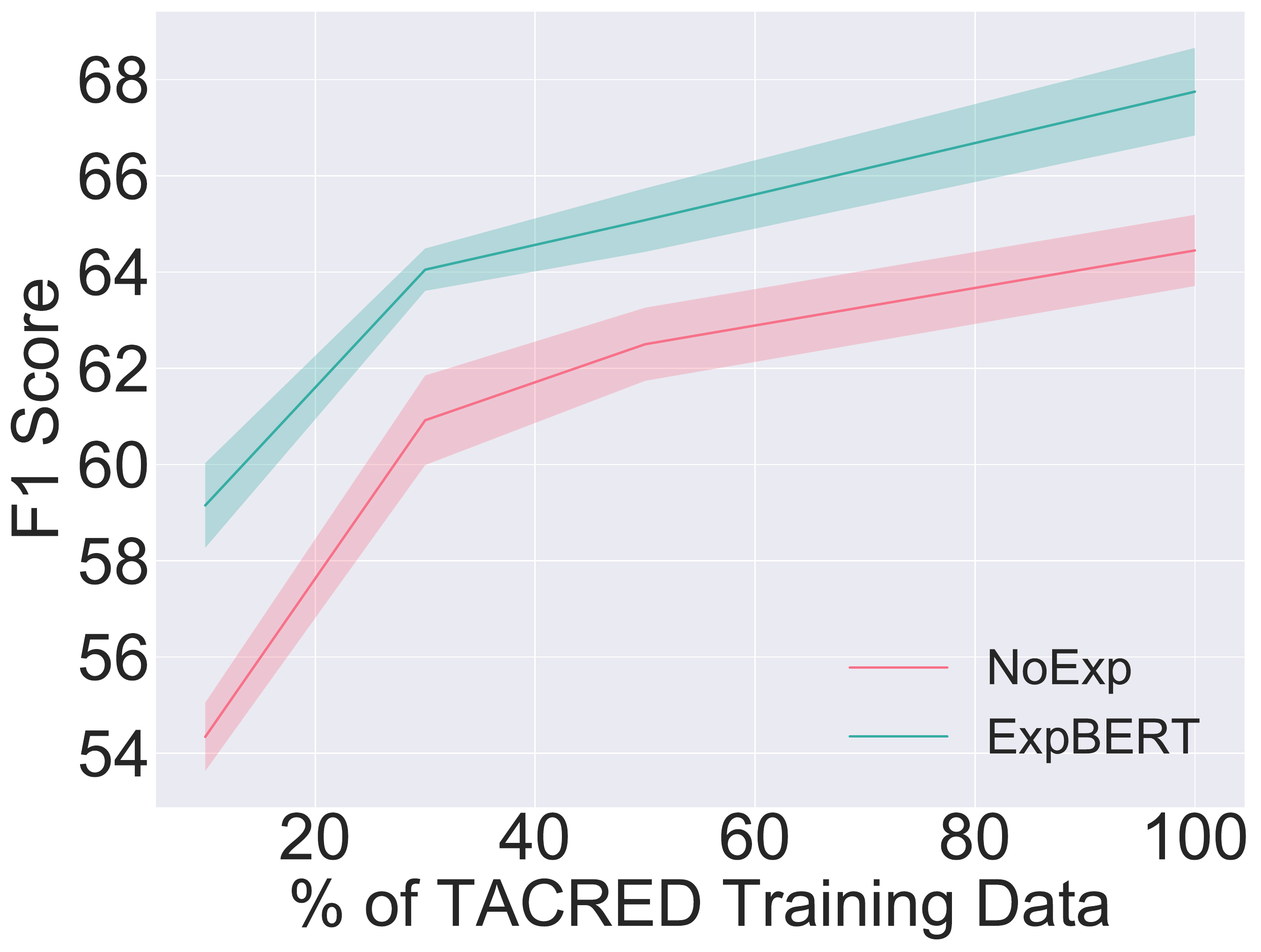}
  \label{fig:sub2}
\end{subfigure}
\caption{$\exbert$ matches the performance of the $\noexp$ baseline with 20x less data on \spouse (Left), and with 3x less data on \tacred (Right).}
\label{fig:data-efficiency}
\end{figure}

\paragraph{Data efficiency.} Collecting a set of explanations $\sE$ requires additional effort---it took the authors about $1$ minute or less to construct each explanation, though we note that it only needs to be done once per dataset (not per example). 
However, collecting a small number of explanations can significantly and disproportionately reduce the number of labeled examples required. 
We trained $\exbert$ and the $\noexp$ baseline with varying fractions of \spouse and \tacred training data (Figure \ref{fig:data-efficiency}).
$\exbert$ matches the $\noexp$ baseline with 20x less data on \spouse; i.e., we obtain the same performance with $\exbert$ with 40 explanations and 2k labeled training examples as with $\noexp$ with 22k examples. On \tacred, $\exbert$ requires 3x less data, obtaining the same performance with 128 explanations and 23k training examples as compared to $\noexp$ with 68k examples. 
These results suggest that the higher-bandwidth signal in language can help models be more data-efficient.

\section{Analysis}
\subsection{Which explanations are important?}
To understand which explanations are important, we group explanations into a few semantic categories (details in \refapp{app-explanations}) and cumulatively add them to the $\noexp$ baseline. In particular, we break down explanations for \spouse into the groups \textsc{Married} (10 explanations), \textsc{Children} (5 explanations), \textsc{Engaged} (3 explanations), \textsc{Negatives} (13 explanations) and \textsc{Misc} (9 explanations). We find that adding new explanation groups helps performance (Table \ref{tab:exp_ablation}), which suggests that a broad coverage of various explanatory factors could be helpful for performance. We also observe that the \textsc{Married} group (which contains paraphrases of \textit{\{$o_1$\} is married to \{$o_2$\}}) alone boosts performance over $\noexp$, which suggests that a variety of paraphrases of the same explanation can improve performance.
\begin{table}[]
\caption{Importance of various explanation groups.}
\label{tab:exp_ablation}
\centering
\small
\begin{tabular}{lr} \toprule
Model           & \spouse   \\ \hline
$\noexp$           & 52.9 $\pm$ 0.97 \\
+ \textsc{Married} & 55.2 $\pm$  0.43  \\
+ \textsc{Children} & 55.9 $\pm$ 0.98  \\
+ \textsc{Engaged} & 57.0 $\pm$ 2.57  \\
+ \textsc{Negatives} & 60.1 $\pm$ 0.87   \\
+ \textsc{Misc} (full $\exbert$)  & 63.5 $\pm$ 1.40 \\ \bottomrule
\end{tabular}
\end{table}
\subsection{Quality vs. quantity of explanations}
We now test whether $\exbert$ can do equally well with the same number of \emph{random} explanations, obtained by replacing words in the explanation with random words. The results are dataset-specific: random explanations help on \spouse but not on \disease. 
However, in both cases, random explanations do significantly worse than the original explanations (Table \ref{tab:random_exp_ablation}).
Separately adding 10 random explanations to our original explanations led to a slight drop ($\approx$1 F1 point) in accuracy. These results suggest that $\exbert$'s performance comes from having a diverse set of high quality explanations and are not just due to providing more features. 

\begin{table}[]
  \caption{$\exbert$ accuracy is significantly lower when we replace words in the original explanations with random words.}
\label{tab:random_exp_ablation}
\centering
\small
\begin{tabular}{lll} \toprule
Model           & \spouse & \disease   \\ \hline
$\noexp$  & 52.9 $\pm$ 0.97 & 49.7 $\pm$ 1.01 \\
$\exbert$ (random) & 56.4 $\pm$ 1.20  &   49.6 $\pm$ 1.22\\
$\exbert$ (orig)  & 63.5 $\pm$ 1.40 & 52.4 $\pm$ 1.23\\ 
$\exbert$ (orig + random) & 62.4 $\pm$ 1.41  & 51.8 $\pm$ 1.03 \\ \bottomrule
\end{tabular}
\end{table}
\subsection{Complementing language explanations with external databases}
Natural language explanations can capture different types of inductive biases and prior knowledge, but some types of prior knowledge are of course better introduced through other means.
We wrap up our experiments with a vignette on how language explanations can complement other forms of feature and representation engineering.
We consider \disease, where we have access to an external ontology (Comparative Toxicogenomic Database or CTD) from \citet{wei2015overview} containing chemical-disease interactions. Following \citet{Hancock2018TrainingCW},
we add 6 bits to the explanation representation $v(x)$ that test if the given chemical-disease pair follows certain relations in CTD (e.g., if they are in the ctd-therapy dictionary).
Table \ref{tab:exbert-combination} shows that as expected, other sources of information can complement language explanations in $\exbert$.

\begin{table}[]
\caption{Combining language explanations with the external CTD ontology improves accuracy on \disease.}
\label{tab:exbert-combination}
\centering
\small
\begin{tabular}{ll} \toprule
Model             & \disease    \\ \hline
$\exbert$           & 52.4 $\pm$ 1.23 \\
$\exbertext$ & 59.1 $\pm$ 3.26  \\ \bottomrule
\end{tabular}
\end{table}

\section{Related work}
Many other works have used language to guide model training. As mentioned above, semantic parsers have been used to convert language explanations into features \citep{srivastava-etal-2017-joint} and noisy labels on unlabeled data \citep{Hancock2018TrainingCW, wang2019learning}.

Rather than using language to define a global collection of features, \citet{rajani-etal-2019-explain} and \citet{esnli18} use \emph{instance}-level explanations to train models that generate their own explanations.
\citet{zaidan-eisner-2008-modeling} 
ask annotators to highlight important words, then learn a generative model over parameters given these rationales. 
Others have also used language to directly produce parameters of a classifier \citep{Ba2015PredictingDZ} and as part of the parameter space of a classifier \citep{Andreas2017LearningWL}. 

While the above works consider learning from static language supervision, \citet{Li2016LearningTD} and \citet{Weston2016DialogbasedLL} learn from language supervision in an interactive setting. In a related line of work, \citet{wang2017naturalizing}, users teach a system high-level concepts via language.

\section{Discussion}
Recent progress in general-purpose language representation models like BERT open up new opportunities to incorporate language into learning. 
In this work, we show how using these models with natural language explanations can allow us to leverage a richer set of explanations than if we were constrained to only use explanations that can be programmatically evaluated, e.g., through n-gram matching ($\bertregex$) or semantic parsing ($\bertsemparse$).

The ability to incorporate prior knowledge of the ``right'' inductive biases into model representations dangles the prospect of building models that are more robust. However, more work will need to be done to make this approach more broadly applicable. We outline two such avenues of future work. First, combining our ExpBERT approach with more complex state-of-the-art models can be conceptually straightforward (e.g., we could swap out BERT-base for a larger model) but can sometimes also require overcoming technical hurdles. For example, we do not fine-tune ExpBERT in this paper; doing so might boost performance, but fine-tuning through all of the explanations on each example is computationally intensive. 

Second, in this paper we provided a proof-of-concept for several relation extraction tasks, relying on the fact that models trained on existing natural language inference datasets (like MultiNLI) could be applied directly to the input sentence and explanation pair. Extending ExpBERT to other natural language tasks where this relationship might not hold is an open problem that would entail finding different ways of interpreting an explanation with respect to the input.

\section*{Acknowledgements}
We are grateful to Robin Jia, Peng Qi, John Hewitt, Amita Kamath, and other members of the Stanford NLP Group for helpful discussions and suggestions. We also thank Yuhao Zhang for assistance with TACRED experiments.
PWK was supported by the Facebook Fellowship Program.
Toyota Research Institute (TRI) provided funds to assist the authors with their research but this article solely reflects the opinions and conclusions of its authors and not TRI or any other Toyota entity.

\section*{Reproducibility}
Code and model checkpoints are available at \href{https://github.com/MurtyShikhar/ExpBERT}{https://github.com/MurtyShikhar/ExpBERT}. The features generated by various interpreters can also be found at that link.

\bibliography{acl2020}
\bibliographystyle{acl_natbib}
\clearpage
\appendix
\section{Appendix}\label{sec:appendix}

\subsection{Implementation Details}\label{sec:app-model}

\paragraph{Interpreting explanations.}
When interpreting an explanation $\expl{i}$ on a particular example $\inp = (\seq, o_1, o_2)$, 
we first substitute $o_1$ and $o_2$ into the placeholders in the explanation $\expl{i}$ to produce an instance-level version of the explanation. For example, ``$\{o_1\}$ and $\{o_2\}$ are a couple'' might become ``Jim Bob and Michelle Duggar are a couple''.

\paragraph{Model hyperparameters and evaluation.}
We use \textsc{bert-base-uncased} for \spouse and \tacred, and \textsc{scibert-scivocab-uncased} for \disease from \citet{Beltagy2019SciBERT}. We finetune all our BERT models on MultiNLI using the Transformers library\footnote{\href{https://huggingface.co/transformers/}{https://huggingface.co/transformers/}} using default parameters. The resulting BERT model is then frozen and used to produce features for our classifier. We use the following hyperparameters for our MLP classifier: number of feed-forward layers $\in$ [0,1], dimension of each layer $\in$ [64, 256], and dropout $\in$ [0.0, 0.3]. We optionally project the 768 dimensional BERT feature vector down to 64 dimensions. To train our classifier, we use the Adam optimizer \citep{Kingma2014AdamAM} with default parameters, and batch size $\in$ [32, 128]. 

We early stop our classifier based on the F1 score on the validation set, and choose the hyperparameters that obtain the best early-stopped F1 score on the validation set. For \spouse and \disease, we report the test F1 means and 95\% confidence intervals of 5-10 runs. For \tacred, we follow \citet{zhang2017tacred}, and report the test F1 of the median validation set F1 of 5 runs corresponding to the chosen 
hyperparameters. 

\subsection{Datasets}\label{sec:app-datasets}
\spouse and \disease preprocessed datasets were obtained directly from the codebase provided by \citet{Hancock2018TrainingCW}\footnote{\href{https://worksheets.codalab.org/worksheets/0x900e7e41deaa4ec5b2fe41dc50594548/}{https://worksheets.codalab.org/worksheets/0x900e7e41deaa4ec5b2fe41dc50594548/}}. We use the train, validation, test split provided by \citet{Hancock2018TrainingCW} for \disease, and split the development set of \spouse randomly into a validation and test set (the split was done at a document level). To process \tacred, we use the default BERT tokenizer and indexing pipeline in the Transformers library.  

\subsection{Explanations}\label{sec:app-explanations}
The explanations can be found in Tables \ref{tab:spouse_exp} and \ref{tab:disease_exp} on the following page. We use 40 explanations for \spouse, 28 explanations for \disease, and 128 explanations for \tacred (in accompanying file). The explanations were written by the authors.

\begin{table}[b]
    \centering
    \small{
    \begin{tabular}{|l|} \hline
\{$o_1$\} and \{$o_2$\} have a marriage license \\
\{$o_1$\}'s husband is \{$o_2$\} \\
\{$o_1$\}'s wife is \{$o_2$\} \\
\{$o_1$\} and \{$o_2$\} are married \\
\{$o_1$\} and \{$o_2$\} are going to tie the knot \\
\{$o_1$\} married \{$o_2$\} \\
\{$o_1$\} and \{$o_2$\} are a married couple \\
\{$o_1$\} and \{$o_2$\} had a wedding \\
\{$o_1$\} and \{$o_2$\} married in the past \\
\{$o_1$\} tied the knot with \{$o_2$\} \\ \hline
\{$o_1$\} and \{$o_2$\} have a son \\
\{$o_1$\} and \{$o_2$\} have a daughter \\
\{$o_1$\} and \{$o_2$\} have kids together \\
\{$o_1$\} and \{$o_2$\} are expecting a son \\
\{$o_1$\} and \{$o_2$\} are expecting a daughter \\ \hline
\{$o_1$\} is engaged to \{$o_2$\} \\
\{$o_1$\} is the fianc\'e of \{$o_2$\} \\
\{$o_1$\} is the fianc\'ee of \{$o_2$\} \\ \hline
\{$o_1$\} is the daughter of \{$o_2$\} \\
\{$o_1$\} is the mother of \{$o_2$\} \\
\{$o_1$\} and \{$o_2$\} are the same person \\
\{$o_1$\} is the same person as \{$o_2$\} \\
\{$o_1$\} is married to someone other than \{$o_2$\} \\
\{$o_1$\} is the father of \{$o_2$\} \\
\{$o_1$\} is the son of \{$o_2$\} \\
\{$o_1$\} is marrying someone other than \{$o_2$\} \\
\{$o_1$\} is the ex-wife of \{$o_2$\} \\
\{$o_1$\} is a location \\
\{$o_2$\} is a location \\ 
\{$o_1$\} is an organization \\
\{$o_2$\} is an organization \\ \hline
\{$o_1$\} and \{$o_2$\} are partners \\
\{$o_1$\} and \{$o_2$\} share a home \\
\{$o_1$\} and \{$o_2$\} are a couple \\
\{$o_1$\} and \{$o_2$\} share the same surname \\
someone is married to \{$o_1$\} \\
someone is married to \{$o_2$\} \\
\{$o_1$\} is a person \\
\{$o_2$\} is a person \\
\{$o_1$\} and \{$o_2$\} are different people \\ \hline
\end{tabular}
}
\caption{Explanations for \spouse. The groups correspond to \textsc{Married}, \textsc{Children}, \textsc{Engaged}, \textsc{Negatives} and \textsc{Misc}. }
\label{tab:spouse_exp}
\end{table}

\begin{table}[h!]
    \centering
    \small{
    \begin{tabular}{|l|}
    \hline
The symptoms of \{$o_2$\} appeared after the \\
administration of \{$o_1$\} \\
\{$o_2$\} developed after \{$o_1$\} \\
Patients developed \{$o_2$\} after being treated with \{$o_1$\} \\
\{$o_1$\} contributes indirectly to \{$o_2$\} \\
\{$o_1$\} has been associated with the development of \{$o_2$\} \\
Symptoms of \{$o_2$\} abated after withdrawal of \{$o_1$\} \\
A greater risk of \{$o_2$\} was found in the \{$o_1$\} group \\
compared to a placebo \\
\{$o_2$\} is a side effect of \{$o_1$\} \\
\{$o_2$\} has been reported to occur with \{$o_1$\} \\
\{$o_2$\} has been demonstrated after the \\
administration of \{$o_1$\} \\
\{$o_1$\} caused the appearance of \{$o_2$\} \\
Use of \{$o_1$\} can lead to \{$o_2$\} \\
\{$o_1$\} can augment \{$o_2$\} \\
\{$o_1$\} can increase the risk of \{$o_2$\} \\
Symptoms of \{$o_2$\} appeared after dosage of \{$o_1$\} \\
\{$o_1$\} is a chemical \\
\{$o_2$\} is a disease \\
\{$o_1$\} is used for the treatment of \{$o_2$\} \\
\{$o_1$\} is known to reduce the symptoms of \{$o_2$\} \\
\{$o_1$\} is used for the prevention of \{$o_2$\} \\
\{$o_1$\} ameliorates \{$o_2$\} \\
\{$o_1$\} induces \{$o_2$\} \\
\{$o_1$\} causes a disease other than \{$o_2$\} \\
\{$o_1$\} is an organ \\
administering \{$o_1$\} causes \{$o_2$\} to worsen \\
\{$o_1$\} is effective for the treatment of \{$o_2$\} \\
\{$o_1$\} has an effect on \{$o_2$\} \\
\{$o_1$\} has an attenuating effect on \{$o_2$\} \\ \hline
    \end{tabular}
    }
    \caption{Explanations for \disease}
    \label{tab:disease_exp}
\end{table}

\end{document}